\documentclass{article}
\usepackage{spconf,amsmath,graphicx}
\usepackage{subcaption,graphicx}

\DeclareMathOperator*{\argmax}{argmax}

\usepackage{algorithm}
\usepackage{algorithmic}
\usepackage[enable]{easy-todo}
\usepackage{amssymb}
\usepackage{arydshln}
\usepackage{pifont}
\usepackage{booktabs}

\usepackage{tikz}
\usetikzlibrary{scopes}
\usetikzlibrary{calc}

\usetikzlibrary{positioning}

\newcommand{\cmark}{\ding{51}}%

\title{Defending Against Physical Adversarial Patch Attacks on Infrared Human Detection}
\name{Lukas Strack$^{1*}$, Futa Waseda$^{2,3*}$, Huy H. Nguyen$^3$, Yinqiang Zheng$^2$, and Isao Echizen$^{2,3}$
\thanks{
This work was partially supported by JSPS KAKENHI Grants JP21H04907 and JP24H00732, and by JST CREST Grants JPMJCR18A6 and JPMJCR20D3, and by JST AIP Acceleration Grant JPMJCR24U3 Japan.
}}
\address{\small{$^{1}$University of Freiburg, Germany  \ \ \ \ \ \ \ \  $^{2}$The University of Tokyo, Japan \ \ \ \ \ \ \ \ $^{3}$National Institute of Informatics, Japan} \\
\small{$^{*}$These authors contributed equally}}

\begin{document}
\ninept

\maketitle

\begin{abstract}
Infrared detection is an emerging technique for safety-critical tasks owing to its remarkable anti-interference capability. However, recent studies have revealed that it is vulnerable to physically-realizable adversarial patches, posing risks in its real-world applications. To address this problem, we are the first to investigate defense strategies against adversarial patch attacks on infrared detection, especially human detection. We propose a straightforward defense strategy, patch-based occlusion-aware detection (POD), which efficiently augments training samples with random patches and subsequently detects them. POD not only robustly detects people but also identifies adversarial patch locations. Surprisingly, while being extremely computationally efficient, POD easily generalizes to state-of-the-art adversarial patch attacks that are unseen during training. Furthermore, POD improves detection precision even in a clean (i.e., no-attack) situation due to the data augmentation effect. Our evaluation demonstrates that POD is robust to adversarial patches of various shapes and sizes. The effectiveness of our baseline approach is shown to be a viable defense mechanism for real-world infrared human detection systems, paving the way for exploring future research directions.
\end{abstract}
\section{Introduction}
\label{sec:intro}

Computer vision models based on deep neural networks (DNNs) exhibit impressive performance across diverse applications.
However, they are vulnerable to adversarial examples~\cite{szegedy2013intriguing, goodfellow2014explaining_harnessing_ae}, i.e., maliciously manipulated inputs designed to deceive DNNs, thereby posing huge risks in real-world applications due to unintended network behaviors.
Moreover, adversarial examples can be implemented in the physical world~\cite{kurakin2018physical_adv, brown2017adv_patch, song2018physical_adv_patch_obj}, such as through the use of adversarial patches~\cite{brown2017adv_patch, eykholt2018adv_patch_cls}.
Unlike adversarial examples that perturb entire images with imperceptible noise~\cite{szegedy2013intriguing, goodfellow2014explaining_harnessing_ae}, adversarial patches alter specific regions of images with salient perturbations.
Thus, these patches can be physically implemented, e.g., printed out, and directly attached to real-world objects to deceive DNN-based computer vision models.

While most of the studies on adversarial patches have focused on RGB-based computer vision~\cite{song2018phys_adv_patch_obj_1, lee2019physical_adv_patch_det, thys2019fooling_adv_patch_human_det}, recent investigations have revealed that infrared object detection models are also vulnerable to physically-realizable adversarial patches~\cite{zhu2021bulb_attack, zhu2022qr_attack, wei2023hotcold, wei2023shape_attack}.
Infrared object detection is an emerging technique for safety-critical applications due to its remarkable anti-interference capability even in harsh environments. It has been applied to a wide range of tasks, such as infrared pedestrian detection~\cite{suard2006pedestrian_infrared}; however, its vulnerability to physical adversarial patches raises substantial concerns regarding its reliability.
Zhu et al.~\cite{zhu2021bulb_attack} showed that a carefully designed physical board with small bulbs could greatly degrade the precision of an infrared human detector. Zhu et al.~\cite{zhu2022qr_attack} designed adversarial ``QR code" pattern clothing that can fool an infrared human detector. Wei et al.~\cite{wei2023hotcold} introduced the HOTCOLD Block (HCB) physical attack against thermal infrared imaging that uses wearable ``warming and cooling pastes" to create infrared adversarial patches with a less conspicuous design. Wei et al.~\cite{wei2023shape_attack} subsequently enhanced the effectiveness of this attack by optimizing the shape and location of the patches. 

However, effective defense strategies against infrared adversarial patches remain fully unexplored.
In fact, the proposed infrared adversarial patches have not been tested against proper defenses specifically designed for adversarial patches, leaving the full extent of the risks unclear.
For example, the HCB attack~\cite{wei2023hotcold} was tested on detectors without any defense, and the Shape-Loc attack~\cite{wei2023shape_attack} was evaluated on defenses for $L_p$-bounded adversarial perturbations, but not defenses designed for adversarial patches.

To this end, we are the first to investigate defense strategies against adversarial patch attacks on infrared detection, focusing on human detection.
We propose a computationally efficient yet effective defense method named patch-based occlusion-aware detection (POD), which efficiently augments training samples with random patches and subsequently detects them. 
These augmented samples simulate occlusions, enabling the model to handle scenarios where human bodies may be obscured or hidden in an attack scenario.
Intriguingly, despite its simplicity, POD demonstrates generalizability to state-of-the-art infrared adversarial patch attacks, even though the patch attacks were unseen during training. Additionally, evaluation using state-of-the-art infrared adversarial patches optimized for shape and location~\cite{wei2023shape_attack} demonstrated that POD is robust to various shapes and sizes of adversarial patches. 
Furthermore, in contrast to typical adversarial training~\cite{goodfellow2014explaining_harnessing_ae}, which sacrifices accuracy in clean (i.e., no-attack) situations, POD instead improves detection precision in clean situations.

Our study highlights that state-of-the-art infrared patch attacks are not as effective as previously believed, as our straightforward data-augmentation-based defense strategy proved highly effective in countering them.
In other words, our findings demonstrate that crafting strong, physically realizable infrared patches remains challenging.
 
The contribution of this work is summarized as follows:
\begin{itemize}
    \item We are the first to investigate defense strategies against physical adversarial patch attacks on infrared detection, opening the door to reducing the vulnerability of infrared detection.
    \item We devise a computationally efficient yet effective defense method, patch-based occlusion-aware detection (POD), which efficiently augments training samples with random patches and subsequently detects them.
    \item POD is easily generalizable to state-of-the-art infrared adversarial patch attacks that are unseen during training, is robust to adversarial patches of various shapes and sizes, and improves detection precision in clean situations.
    \item We highlight that state-of-the-art infrared patch attacks are not as effective as previously believed. Thus, our work encourages future research to evaluate infrared patch attacks against proper defenses, promoting a better understanding of the associated risks in realistic scenarios.
\end{itemize}




\section{Physical Adversarial Patches}
\label{sec:attack}

In this section, we first explain the general objective of generating adversarial patches. Then, we describe the constraints when crafting physical adversarial patches that interfere with infrared human detection.

\subsection{General Framework}
Let $f_{\theta}$ represent the object detector parameterized by $\theta$, and $f_{\theta}(x)$ be the predicted output for an infrared image $x \in \mathbb{R}^{h \times w}$.
We define $A(p, x)$ as the operator for applying adversarial patch $p$ to image $x$ (assuming a fixed patch location for simplicity).

Here, the attacker aims to optimize a single ``universal" adversarial patch that is scene-agnostic, intending to deceive the human detector $f_{\theta}$, irrespective of the pose or image context.
The objective in generating an adversarial patch can thus be formulated as,
\begin{eqnarray}
     \argmax_p \mathbf{E}_{(x,y) \sim D} \left[ J\left(f_{\theta}(A(p,x)), y \right) \right]
\end{eqnarray}
where $(x,y)$ is an input-label pair from the data distribution $D$, and $J\left(\cdot, \cdot \right)$ is an arbitrary loss function that quantifies the discrepancy between the ground truth and the prediction.
The core concept is to optimize a single patch $p$ that maximizes the loss function across the entire data distribution, inducing inaccurate object detection.

More specifically, when attacking an object detector, the loss function $J\left(\cdot, \cdot \right)$ can be computed for either ``objectness" or ``classification" scores of an object detector; The object detector outputs an objectness score (indicating the likelihood of a bounding box containing an object) and a classification score (identifying the object's class within the bounding box). Thys et al.~\cite{thys2019fooling_adv_patch_human_det} has shown that minimizing the objectness score is more effective than inducing misclassification, and the state-of-the-art infrared adversarial patches, such as HCB attack~\cite{wei2023hotcold} and the Shape-Loc attack~\cite{wei2023shape_attack}, follows the approach.

\subsection{Constraints}
\label{subsec:constraints}
In addition to the above objective, it is essential to consider constraints during the optimization process, particularly for infrared adversarial patches in physical space.
There are two key constraints to be considered in the infrared scenario: (1) infrared images have much less texture information than RGB images due to overlap in the spectral reflectance across different materials; (2) physically implementing a high-resolution patch or one with a pixel value of a specific magnitude is challenging due to material limitations.

Constraints are thus added during optimization to obtain physically-realizable patches for interfering with infrared detection~\cite{zhu2021bulb_attack, zhu2022qr_attack, wei2023hotcold, wei2023shape_attack}. 
For example, with the HCB attack~\cite{wei2023hotcold}, only nine grids are used for a patch with binary values (0/1). Wei et al.~\cite{wei2023shape_attack} optimized patches with only binary values while promoting the clustering of pixels with the same value for easy real-world implementation.

This limitation of low-resolution and simple-colored infrared adversarial patches raises the question, ``Are state-of-the-art infrared patches effective against properly defended models?".
This motivated our investigation into the straightforward defense mechanism POD, detailed in the next section.

\section{Patch-based Occlusion-aware Detection}
\label{sec:method}

\begin{figure*}[t]
\centering  
\includegraphics[width=\linewidth]{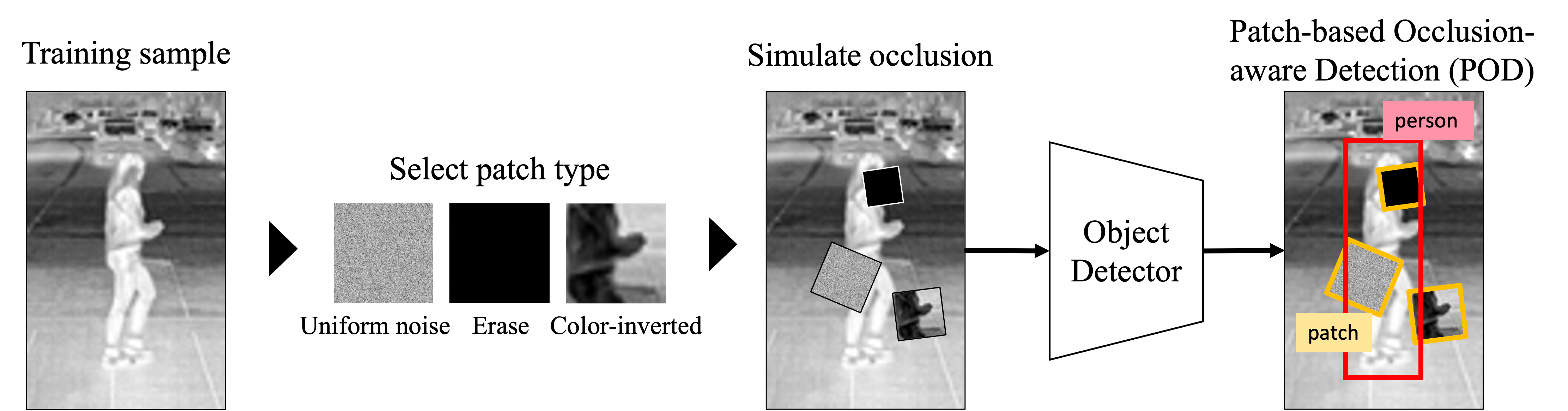}
\caption{Training pipeline of Patch-based Occlusion-aware Detection (POD). 
POD efficiently augments training samples with random patches and subsequently detects them. The augmented samples
simulate occlusions, enabling the model to handle scenarios where
human bodies may be obscured or hidden in attack scenarios.}
\label{fig:pod_pipeline}
\end{figure*}

Our proposed defense mechanism, POD, is simple yet highly effective against infrared adversarial patch attacks.
Our key idea involves augmenting training samples with random patches and training an object detection model. 
We introduce a "patch" class in addition to the "human" class for the model to detect these patches.

POD has three key characteristics:
\begin{itemize}
    \item{\textbf{Generalization without attack algorithm assumption:} The POD model is robust against sophisticated adversarial patches unseen during training despite the use of simple augmentation techniques.}
    \item{\textbf{Robust human detection:} POD not only robustly detects people in the presence of adversarial patches but also is trained to identify them. This results in improved generalization against previously unseen adversarial patches.}
    \item{\textbf{Simple quick training:} The training time is much less than that of traditional adversarial training schemes~\cite{goodfellow2014explaining}, which aims to solve the min-max problem by directly feeding adversarial examples during training.}
\end{itemize}
The overall pipeline is illustrated in Fig.~\ref{fig:pod_pipeline}.

\subsection{Adding Random Patch-based Occlusions}
\label{sec:rand-patch-occlusion}

We introduce a simple training strategy to train a robust human detection model, i.e., apply random patches to simulate occlusions, which can obscure or hide people. 
Unlike conventional adversarial training schemes~\cite{goodfellow2014explaining, ji2021adv_yolo} that augment training samples using specific attack algorithms, we train our detection model across simple yet diverse occlusion scenarios, resulting in general resilience against unforeseen adversarial patches. 
Our strategy harnesses the power of simplicity and randomness, inspired by TrivialAugment strategy~\cite{müller2021trivialaugment}, which achieved state-of-the-art performances by simplifying over-complicated augmentations for image classification pipelines. Our approach is highly inspired by this approach, realizing a simple yet powerful defense based on randomness.
Additionally, considering that physical infrared adversarial patches typically exhibit basic textures and simple colors, as explained in Sec.~\ref{subsec:constraints}, it is expected that a simple and efficient defense method is particularly effective in defending against them.

The patches for our training strategy are designed to exhibit random characteristics, such as size, shape, texture, and placement, to simulate realistic scenarios in which adversarial patches might be encountered.
The patches are created by first cropping a square region from an original image, with their size, rotation degree, and location being randomly determined. 
Next, the random patch is added to the cropped region to simulate an adversarial patch.
We use three random patch variations:
\begin{itemize}
    \item{\textbf{Erase patch $p_{erase}$} simulates the most basic occlusion that hides persons. The patch has the pixel values of zero.}
    \item{\textbf{Color-inverted patch $p_{inv}$} mimics an infrared adversarial patch with a more sophisticated texture. Patches with texture complexity similar to those captured by infrared cameras are efficiently mimicked through color inversion.}
    \item{\textbf{Uniform noise patch $p_{noise}$} mimics an infrared adversarial patch with an intricate texture, although physical implementation remains challenging. It applies a random value sampled from a uniform distribution to each pixel.}
\end{itemize}

\begin{figure*}[t]
\centering  
\includegraphics[width=\linewidth]{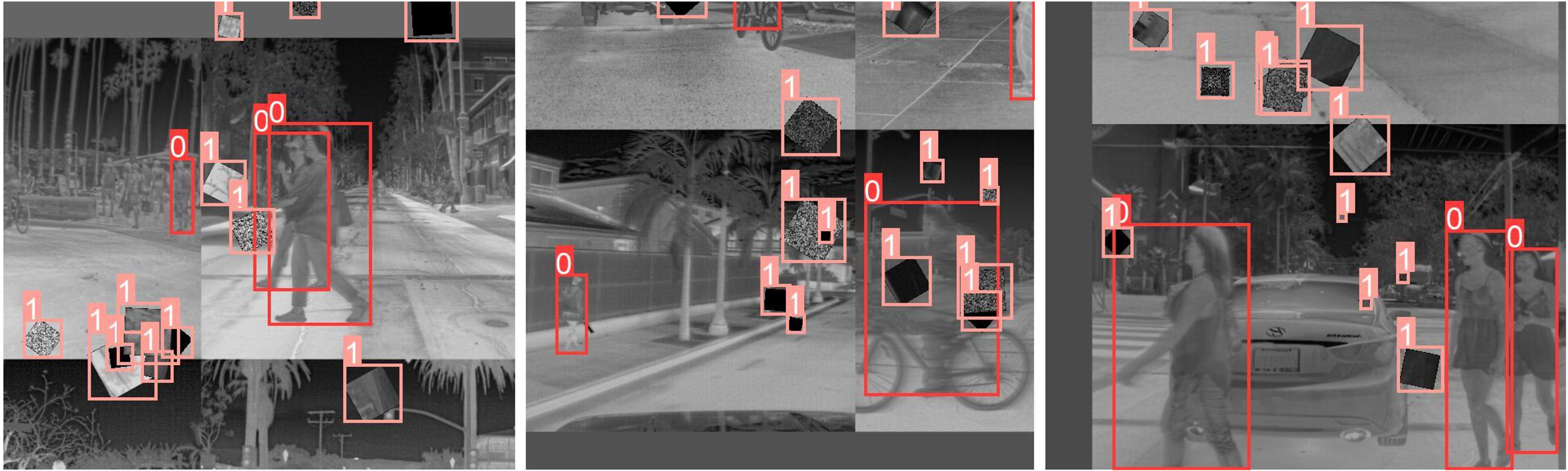}
\caption{Example training samples for patch-based occlusion-aware detection (POD). Labels ``0" and ``1" signify ``human" and ``patch" classes, respectively. POD aims not only to detect people with random occlusions but also to identify patch locations.}
\label{fig:pod_training_samples}
\end{figure*}

The entire augmentation process is outlined in Algorithm~\ref{PODalgo}. 
The examples of augmented training samples are shown in Fig.~\ref{fig:pod_training_samples}.

\begin{algorithm}
\textbf{Input:} Image $I$ with width $w$ and height $h$  \\
\textbf{Output:} Augmented image $I'$ \\
\textbf{Initialization:} $I' \leftarrow I$ , $N \leftarrow$  Random number of patches

\begin{algorithmic}[1]
\FOR {i in N}
    \STATE // Determine a patch
    \STATE $p \leftarrow$ Select patch type from $\{p_{erase}, p_{inv}, p_{noise}\}$.
    \STATE $l \leftarrow$  Random patch size.
    \STATE $\theta \leftarrow$ Random rotation degree of patch.
    \STATE $c_x, c_y \leftarrow$ Random center position of patch. 
    \STATE
    \STATE // Apply the patch to the image
    \STATE $I' \leftarrow$ apply patch $p^{l\times l}$ with rotation degree $\theta$ at $(c_x, c_y)$.
\ENDFOR
\end{algorithmic}
\caption{POD Training Augmentation Process}
\label{PODalgo}
\end{algorithm}

\subsection{Detecting Patch-based Occlusions}
\label{subsec:det_patch}

POD involves training the model not only to detect people but also to explicitly identify the existence and location of patch-based occlusions present in an image. 
The model is thus able to handle the uncertainties introduced by occlusions, which can be crucial for reliable decision-making in practical applications.
Intriguingly, we found that explicitly training the model to identify the locations of patch-based occlusions noticeably improved its robustness against strong adversarial patches, as explained in a later section.

To detect adversarial patches, we modify an object detection model to have an additional patch class.
Here, the types of patches are not distinguished, and only one extra class is used to detect patch-based occlusions. 

Additionally, since the primary objective is to detect people robustly, we employ weighted cross-entropy loss~\cite{phan2020resolving_wce} to prioritize the human class over the patch class:
\begin{equation}
    \alpha \cdot L^{human}_{CE} + (1 -\alpha) \cdot L^{patch}_{CE}
\end{equation}
where $L^{human}_{CE}$ and $L^{patch}_{CE}$ represent the classification loss with cross-entropy for humans and patches, respectively.
This approach helps address the class imbalance between humans and patches after POD augmentation, enabling the model to prioritize and concentrate on human detection.
In our experiments, the weight for the human class, denoted as $\alpha$, was fixed at 0.9.
Humans and patches are labeled as in Fig.~\ref{fig:pod_training_samples}.


\section{Experimental Setup}
\label{sec:exp}


\subsection{Dataset}
\label{subsec:data}
Following Zhu et al.~\cite{zhu2021bulb_attack} and Wei et al.~\cite{wei2023hotcold}, we evaluated model performance on the Teledyne FLIR ADAS Thermal dataset~\cite{FLIR}. 
Infrared images were acquired with a Teledyne FLIR Tau2 (13 mm f/1.0 with a 45-degree HFOV and 37-degree VFOV), and the thermal camera operated in T-linear mode. 
Following Wei et al.~\cite{wei2023hotcold}, we filtered the original dataset for the task of human detection. First, we used only the ``person" category, excluding images without the person class. Next, we filtered the persons to ensure that the height of their bodies are larger than 120 pixels. This process resulted in 1169 training images with 1810 person labels and 84 test images with 154 person labels.

To evaluate the model against shape-location-optimized (``shape-loc") attacks~\cite{wei2023shape_attack}, we created a custom dataset based on the CVC-14 dataset~\cite{s16060820}, in which each image includes only one person. 
The CVC-14 dataset provides already cropped images of persons. 
From this dataset, we randomly chose 97 images and manually labeled them. 
The custom dataset, along with its annotations, is included in our provided codes.

\subsection{Evaluation}

We used average precision (AP@0.5) to evaluate the ability of object detection models to detect humans (we repeated each experiment five times and reported the standard deviation).

In our experiments, the physical adversarial patches were simulated in digital space. Since real-world attacks are less successful due to changes in lighting and human pose, as well as limitations on physical materials for crafting attacks, digital attacks are the upper bound for attack capability. 
To evaluate the robustness of the evaluated infrared detection models, we compared them in the following scenarios:
\begin{itemize}
    \item \textbf{Random noise patch}: an adversarial patch in which each pixel value is sampled from a uniform distribution. This is \textit{a digital-space attack} that simulates random occlusions.
    \item \textbf{AdvPatch}~\cite{thys2019fooling_adv_patch_human_det}: a universal adversarial patch designed to deceive RGB-based human detectors in the physical world. Due to its complex texture, it cannot be physically implemented in the infrared scenario; we regard this attack as a \textit{strong digital-space attack}.
    \item \textbf{HCB}~\cite{wei2023hotcold}: a state-of-the-art physically-realizable universal adversarial patch utilizing wearable warming and cooling pastes for attacks on infrared detectors.
\end{itemize}


Additionally, we evaluated the Shape-Loc attack~\cite{wei2023shape_attack} to evaluate the model’s performance against adversarial patches of different sizes and shapes. 
These patches are image-dependent, resulting in 97 individual optimized shapes and locations for our custom dataset, described in Sec.~\ref{subsec:data}.


\subsection{Implementation details}
We used the YOLO-v5 model architecture ~\cite{glenn_jocher_2021_5563715_yolo_v5} since models based on this architecture are fast and widely used as detectors. 
For infrared detection, we used the pre-trained weights on the RGB images from the MS COCO (Microsoft Common Objects in Context) dataset~\cite{lin2014microsoft} as the initial weights, then fine-tuned the weights on infrared images from the FLIR ADAS Dataset.
All of the evaluated detection models were trained for 50 epochs for fair comparison. Since infrared images have only one channel, we expanded their channel size to three dimensions to leverage the RGB-pre-trained weights. 

\subsection{Comparison between Adversarial Training Scheme}
Furthermore, we developed a POD variant called ``Adv-POD," incorporating an adversarial training scheme that explicitly inputs adversarial patches during training, modified from Ad-YOLO~\cite{ji2021adv_yolo}.
Unlike Ad-YOLO, which is inadequate for defending against location-optimized patches like HCB and shape-loc due to its central patch placement, Adv-POD employs the same random patch placement scheme as POD. 
This modification allows us to fairly compare the effects of using basic patches (Sec.~\ref{sec:rand-patch-occlusion}) or adversarial patches.

The procedure for Adv-POD comprises four steps: (1) a model is trained using a history of adversarial patches; (2) a new adversarial patch is generated every 15 epochs, starting from the $5_{th}$ epoch; (3) the adversarial patches are stored in the adversarial patch history for subsequent training; (4) the stored patches are randomly selected and attached to input image to train the detection model. We used the AdvPatch attack~\cite{thys2019fooling_adv_patch_human_det} for adversarial patch generation, positioning the patches in the exact same way as for POD, i.e., randomly determining their sizes and locations for fair comparison.

\section{Results}

In this section, we present experimental results demonstrating the effectiveness of POD against diverse adversarial patch attacks. 
The POD's effectiveness is summarized in Table~\ref{tab:conclusion}.

\begin{table*}[htp!]
\centering
\begin{tabular}{lcc|lrrl|l}
\hline
\multicolumn{3}{c|}{Method} & \multicolumn{4}{c|}{Average Precision $\uparrow$} & Training\\
\multicolumn{1}{c}{} & \multicolumn{1}{l}{Patch Type} & \multicolumn{1}{l|}{Patch Detect.} & \multicolumn{1}{c:}{Clean} & \multicolumn{1}{c}{$^\dagger$Noise Patch} & \multicolumn{1}{c:}{$^\dagger$AdvPatch} & \multicolumn{1}{c|}{HCB} & \multicolumn{1}{c}{Time (rel.)} \\ \hline
Std. Training     &  -    & -          & \multicolumn{1}{c:}{.9199 $\pm$ .0064} &        .6436 $\pm$ .0519 &       \multicolumn{1}{c:}{.2066 $\pm$ .0465} & .3616 $\pm$ .1379 & $\times$ 1 \\ \hdashline
POD$_{noDet}$     & Rand. &            & \multicolumn{1}{c:}{.9193 $\pm$ .0164} & $\ast$ .8410 $\pm$ .0207	 &       \multicolumn{1}{c:}{.5091 $\pm$ .0472} & .6582 $\pm$ .0649 & $\times$ 1.0 \\
POD               & Rand. & \cmark & \multicolumn{1}{c:}{.9360 $\pm$ .0082} & $\ast$ .8078 $\pm$ .0170 &       \multicolumn{1}{c:}{.7897 $\pm$ .0239} & .7638 $\pm$ .0374 & $\times$ 1.0 \\
Adv-POD$_{noDet}$ & Adv.  &            & \multicolumn{1}{c:}{.9260 $\pm$ .0149} &        .8744 $\pm$ .0330	 & \multicolumn{1}{c:}{$\ast$.8133 $\pm$ .0607} & .7884 $\pm$ .0372 & $\times$ 5.0 \\
Adv-POD           & Adv.  & \cmark & \multicolumn{1}{c:}{.9395 $\pm$ .0116} &        .8162 $\pm$ .0249 & \multicolumn{1}{c:}{$\ast$.7637 $\pm$ .0120} & .7426 $\pm$ .0428 & $\times$ 5.0
\\ \hline
\end{tabular}
\caption{Average precision (AP@0.5) for adversarial patches. 
Model names with ``$_{noDet}$" indicate models without the patch-detection module described in Sec.~\ref{subsec:det_patch}.
Attacks marked with $\dagger$ are not physically realizable in the infrared scenario.
Results marked with $\ast$ are the results for seen attacks: the adversarial patch was encountered during training.
The rightmost column illustrates the relative training time compared to standard training.
The results highlight the effectiveness of our simple defense strategy, POD, in defending against previously unseen adversarial patches, with negligible additional training time.
Notably, POD enhances detection precision in clean scenarios.
}
\label{tab:results}
\end{table*}


\subsection{Evaluation for Digital and Physical Adversarial Patches}


\textbf{POD is robust to state-of-the-art physically realizable infrared adversarial patch.}
Table~\ref{tab:results} shows that all POD variants are robust to the state-of-the-art physical infrared adversarial patch generated by HCB.
Notably, the Average Precision on HCB is over 70\%, which is only a few percent worse than the random-noise patch scenario; this suggests that the adversarial effect of HCB on the POD model is quite low and the prediction errors induced by HCB predominantly arise from simply concealing the human body.

\textbf{POD generalizes to strong digital-space adversarial patch.}
Table~\ref{tab:results} shows that POD is noticeably robust against the strong digital-space attack, i.e., AdvPatch~\cite{thys2019fooling_adv_patch_human_det}, which was unseen during training.
Given that the digital-space adversarial patches disregard material and sensor constraints for the infrared detection scenarios, as described in Sec~\ref{subsec:constraints}, they are generally stronger than physical-space adversarial patches.
Therefore, despite its simplicity, the POD strategy is suggested to generalize to unforeseen advanced infrared patches with intricate textures and colors.

Additionally, despite POD not being exposed to AdvPatch during training, it achieves a comparable Average Precision against AdvPatch as Adv-POD, a model explicitly trained with AdvPatch.
This may be attributed to Adv-POD suffering from overfitting and failing to generalize to the test samples. 
It is well-known that typical adversarial training suffers from adversarial overfitting~\cite{rice2020overfitting}, wherein the model overfits to worst-case adversarial images within the training samples.
In contrast, POD avoids overfitting due to its simplicity and randomness of augmentations, resulting in generalization against unseen strong attacks.

An important property of POD is that the patch detection module is necessary to defend against AdvPatch; this indicates that explicitly identifying the location of patches during training helps a detector generalize against strong, unforeseen adversarial patches.

\textbf{POD greatly enhanced robustness against random occlusions.}
Interestingly, we observed that a random-noise patch attack is already effective against the standard model due to simply concealing the person’s body, leading to a degradation in Average Precision from 92\% to 64\%.
In contrast, all of the POD variants had an Average Precision of over 80\%. 
This highlights POD's effectiveness in improving resistance to random occlusions that can occur naturally without the presence of attackers.


\textbf{POD achieved performance comparable to that of its adversarial variant, Adv-POD, yet with a notably shorter training time.}
The results in Table~\ref{tab:results} demonstrate that POD and Adv-POD$_{noDet}$ achieved similar performance in all scenarios, although the training time of POD is significantly shorter than that of Adv-POD$_{noDet}$. This is because, with the adversarial training scheme, it is time-consuming to generate adversarial examples during training, whereas POD relies solely on efficient augmentation techniques to simulate patch-based occlusions. 
Therefore, POD is an efficient and practical defense mechanism that can easily scale to train a model on a large dataset.

\textbf{POD improved precision in the clean scenario.}
Notably, the POD variants did not sacrifice detection precision in the clean scenario but rather improved it, in contrast to conventional adversarial training schemes~\cite{goodfellow2014explaining_harnessing_ae, madry2017towards} that often suffer from a trade-off between clean and robust accuracy~\cite{zhang2019theoretically_trade_off}.
This is attributed to the POD variants simulating various patch-based occlusions with diverse patch types with random size and location. 
We presume that the POD variants learn diverse features that are useful to robustly detect humans even when part of the body is occluded, and consequently, they improve accuracy in the clean scenario.

\textbf{Adv-POD with the patch detection module overfit to a specific attack algorithm seen during training.}
While the patch detection module (refer to Sec.~\ref{subsec:det_patch}) was beneficial for the POD strategy, its use in Adv-POD, the adversarial variant of POD, led to a degradation in defending against adversarial patches.
We hypothesize that this is due to overfitting in Adv-POD: a model can overfit when trained on a single attack that lacks patch diversity, and overfitting seems more likely to occur when the model is trained to detect the patch location explicitly.

\subsection{Robustness to Varied Adversarial Patch Sizes and Shapes}
We evaluated the models on the Shape-Loc attack~\cite{wei2023shape_attack}, which uses shape-location-optimized infrared adversarial patches, to understand the model's performance against patches with various sizes and shapes. 
Unlike universal adversarial patches, such as HCB, the Shape-Loc attack generates an image-dependent shape-location-optimized adversarial patch. 
We used the custom dataset based on the CVC-14 dataset~\cite{s16060820} described in Sec.~\ref{subsec:data}.

\begin{figure*}[h]
    \centering
    \begin{subfigure}{.82\columnwidth}
        \centering
        \includegraphics[height=4.9cm]{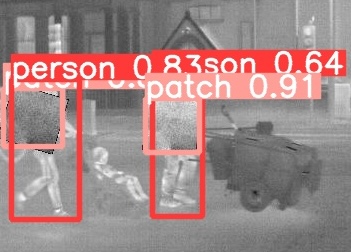}  
        \caption{AdvPatch~\cite{thys2019fooling_adv_patch_human_det}}
        \label{SUBFIGURE LABEL 1}
    \end{subfigure}
    \begin{subfigure}{.58\columnwidth}
        \centering
        \includegraphics[height=4.9cm]{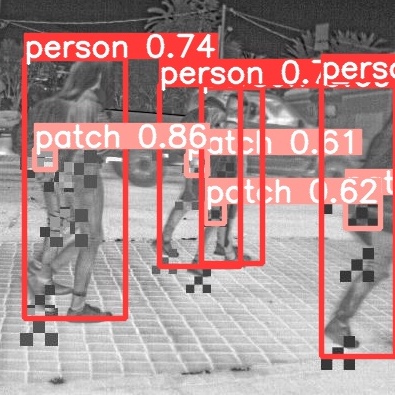}  
        \caption{HCB~\cite{wei2023hotcold}}
        \label{SUBFIGURE LABEL 2}
    \end{subfigure}
    \begin{subfigure}{.58\columnwidth}
        \centering
        \includegraphics[height=4.9cm]{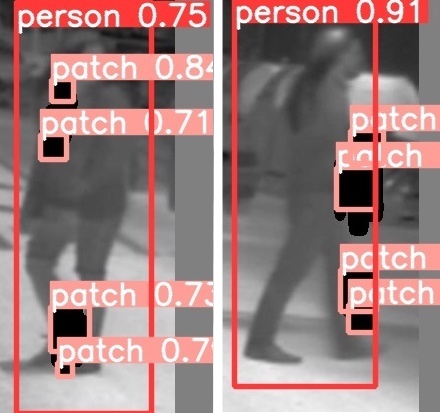}  
        \caption{Shape-Loc~\cite{wei2023shape_attack}}
        \label{SUBFIGURE LABEL 3}
    \end{subfigure}
    \caption{Examples of POD detecting persons and patches for AdvPatch (left), HCB (middle), and shape-loc (right) attacks.}
    \label{fig:detect}
\end{figure*}

The results in Table~\ref{tab:shape_attack} show that POD was robust to the Shape-Loc attack. Its performance is visualized in Fig.~\ref{fig:detect}, which shows that POD effectively identified both people and Shape-Loc attack patches. 
This highlights POD's resilience to adversarial patches with diverse sizes and shapes, even when the shapes and locations are adversarially optimized. 
Hence, our strategy of simulating random patches during training is a reasonably effective and practical approach for mitigating physical infrared adversarial patch attacks.

\begin{table}[H]
\centering
\begin{tabular}{l|l|l}
\hline
\multicolumn{1}{c}{}  & \multicolumn{1}{|c}{Clean} & \multicolumn{1}{|c}{Shape-Loc Attack} \\
\hline
Std. Training          & .9818 $\pm$ .0071 & .5005 $\pm$ .1573 \\ \hdashline
POD$_{noDet}$          & .9845 $\pm$ .0043 & .9346 $\pm$ .0114 \\
POD                    & .9813 $\pm$ .0059	 & .9133 $\pm$ .0318 \\ 
Adv-POD$_{noDet}$      & .9799 $\pm$ .0053 & .9353 $\pm$ .0208 \\
Adv-POD                & .9790 $\pm$ .0059	 & .9090 $\pm$ .0322 \\ \hline
\end{tabular}
\caption{Average precision for adversarial patches with various sizes and shapes used in the Shape-Loc attack~\cite{wei2023shape_attack}.
All POD variants demonstrated improved robustness against Shape-Loc attacks, along with improved precision in clean scenarios.}
\label{tab:shape_attack}
\end{table}

\begin{table}[ht!]
\centering
\scalebox{1.0}{ 
\begin{tabular}{l|cccc}
\hline
\multicolumn{1}{c}{}  & \multicolumn{3}{c}{Defense Capability} & \multicolumn{1}{c}{Training Efficiency} \\
\multicolumn{1}{c}{}  & \multicolumn{1}{c}{Adv.} & \multicolumn{1}{c}{HCB} & \multicolumn{1}{c}{Shape} & \multicolumn{1}{c}{} \\
\hline
Std. Training         &   &  &  & \cmark \\ \hdashline
POD$_{noDet}$         & $\triangle$ & \cmark & \cmark & \cmark \\
POD                   & \cmark & \cmark & \cmark & \cmark \\ 
Adv-POD$_{noDet}$     & \cmark & \cmark & \cmark &  \\
Adv-POD               & \cmark & \cmark  & \cmark & \\ 
\hline
\end{tabular}
}
\caption{Summary of defense capabilities of POD variants. POD, in the 3rd row, demonstrated the best defense capability along with efficient training.}
\label{tab:conclusion}
\end{table}

\subsection{Ablation Study: Patch Augmentation Type in POD}

\begin{table*}[ht]
\centering
\begin{tabular}{c|ccc|ccc:c}
\hline
Method & \multicolumn{3}{c|}{Patch Augmentation Type} & \multicolumn{4}{c}{Average Precision $\uparrow$}\\
& Erase & Color-Inverted & Uniform Noise & Noise Patch & AdvPatch & HCB & Shape-Loc \\
\hline
Std. Training &  &  &  & .6436 $\pm$ .0519 & .2066 $\pm$ .0465 & .3616 $\pm$ .1379 & .5005 $\pm$ .1573 \\ \hdashline
& \cmark &  &  & .8142 $\pm$ .0269 & .4113 $\pm$ .0332 & .6919 $\pm$ .0436 & .8989 $\pm$ .0193 \\
& & \cmark &  & .7931 $\pm$ .0137 & .7796 $\pm$ .0235 & .7858 $\pm$ .0515 & .9240 $\pm$ .0208 \\
 & &  & \cmark & .8191 $\pm$ .0150 & .7750 $\pm$ .0332 & .7287 $\pm$ .0304 & .9262 $\pm$ .0267 \\
POD & \cmark & \cmark &  & .7878 $\pm$ .0191 & .7736 $\pm$ .0265 & .6898 $\pm$ .0593 & .9090 $\pm$ .0322 \\
 & \cmark &  & \cmark & .8063 $\pm$ .0253 & .7898 $\pm$ .0465 & .6694 $\pm$ .0281 & .8918 $\pm$ .0324 \\
& & \cmark & \cmark & .7990 $\pm$ .0162 & .7812 $\pm$ .0210 & .7851 $\pm$ .0141 & .9181 $\pm$ .0312 \\
(default) & \cmark & \cmark & \cmark & .8078 $\pm$ .0170 & .7897 $\pm$ .0239 & .7638 $\pm$ .0374 & .9133 $\pm$ .0318 \\
\hline
\end{tabular}
\caption{Ablation study for patch type used in POD training. We assess Shape-Loc using the CVC-14 dataset, and conduct other attacks on the FLIR ADAS dataset. The results indicate that using all three types of patches results in good AP for various patch attacks.
}
\label{tab:ablation}
\end{table*}

In this section, we conduct an ablation study of the patch augmentation types used in POD training.
As explained in Sec.~\ref{sec:rand-patch-occlusion}, POD simulates three types of occlusions during training: erase patch, color-inverted patch, and uniform noise patch.
These three patches are intended to play different roles in simulating occlusions by having different characteristics, such as texture complexity.

Table~\ref{tab:ablation} results indicate that employing all three patches in POD yields high average precision (AP) across various patch attacks, without conflicts in utilizing multiple patch types.

One finding is that using only the erase patch, which has constant pixel values of 0, led to lower AP compared to other POD variants when tested against state-of-the-art adversarial patches of AdvPatch, HCB, and Shape-Loc attack.
We hypothesize that simulating occlusions with varied texture complexity is crucial to defend against adversarial patches, and the erase patches lacked this diversity.

Nevertheless, we observed that all variants of POD in Table~\ref{tab:ablation} resulted in much better defense capability than the standard training model. 
The APs against the state-of-the-art physically realizable infrared adversarial patch attacks are quite high: all POD variants achieve around 70\% for the HCB attack and around 90\% for the Shape-Loc attack.
This highlights our findings that the state-of-the-art infrared adversarial patch attacks are not as effective as previously believed, since our simple augmentation-based defense strategy greatly improves AP without increasing training time and without sacrificing precision in clean scenarios (even improves).

We believe that further efforts to optimize the augmentation strategy with careful design can improve the performance of POD.
Nevertheless, we would also like to emphasize that POD is designed to be extremely computationally efficient by utilizing only efficiently created patch types, and creating much more sophisticated patch-based occlusions may increase the training time.

\section{Conclusion}
\label{sec:conclusion}

In this work, we were the first to investigate defense strategies against physical adversarial patch attacks in infrared detection, focusing on human detection.
We introduced a computationally efficient yet effective defense strategy, patch-based occlusion-aware detection (POD), which augments the training samples with random patches to simulate diverse occlusions and subsequently detects them.
Interestingly, state-of-the-art infrared patch attacks were much less effective against our simple defense strategy, POD, challenging their perceived defense capability.
POD demonstrated remarkable efficacy in countering infrared adversarial patches unseen during training, and exhibited robustness against patch attacks with varying shapes and sizes.
Furthermore, POD improved detection precision in clean (i.e., no-attack) scenarios by learning diverse features to handle various occlusions.

In summary, our pioneering work highlights that state-of-the-art infrared patch attacks are not as effective as previously believed.
We encourage future research to evaluate infrared patch attacks against proper defenses to understand the associated risks better.

\vfill\pagebreak

\bibliographystyle{IEEEbib}
\bibliography{strings}

\begin{thebibliography}{10}

\bibitem{szegedy2013intriguing}
Christian Szegedy, Wojciech Zaremba, Ilya Sutskever, Joan Bruna, Dumitru Erhan, Ian Goodfellow, and Rob Fergus,
\newblock ``Intriguing properties of neural networks,''
\newblock in {\em ICLR}, 2014.

\bibitem{goodfellow2014explaining_harnessing_ae}
Ian~J Goodfellow, Jonathon Shlens, and Christian Szegedy,
\newblock ``Explaining and harnessing adversarial examples,''
\newblock in {\em ICLR}, 2015.

\bibitem{kurakin2018physical_adv}
Alexey Kurakin, Ian~J Goodfellow, and Samy Bengio,
\newblock ``Adversarial examples in the physical world,''
\newblock in {\em Artificial intelligence safety and security}, pp. 99--112. Chapman and Hall/CRC, 2018.

\bibitem{brown2017adv_patch}
Tom~B Brown, Dandelion Man{\'e}, Aurko Roy, Mart{\'\i}n Abadi, and Justin Gilmer,
\newblock ``Adversarial patch,''
\newblock {\em arXiv preprint arXiv:1712.09665}, 2017.

\bibitem{song2018physical_adv_patch_obj}
Dawn Song, Kevin Eykholt, Ivan Evtimov, Earlence Fernandes, Bo~Li, Amir Rahmati, Florian Tramer, Atul Prakash, and Tadayoshi Kohno,
\newblock ``Physical adversarial examples for object detectors,''
\newblock in {\em 12th USENIX workshop on offensive technologies (WOOT 18)}, 2018.

\bibitem{eykholt2018adv_patch_cls}
Kevin Eykholt, Ivan Evtimov, Earlence Fernandes, Bo~Li, Amir Rahmati, Chaowei Xiao, Atul Prakash, Tadayoshi Kohno, and Dawn Song,
\newblock ``Robust physical-world attacks on deep learning visual classification,''
\newblock in {\em Proceedings of the IEEE conference on computer vision and pattern recognition}, 2018, pp. 1625--1634.

\bibitem{song2018phys_adv_patch_obj_1}
Kevin Eykholt, Ivan Evtimov, Earlence Fernandes, Bo~Li, Amir Rahmati, Florian Tramer, Atul Prakash, Tadayoshi Kohno, and Dawn Song,
\newblock ``Physical adversarial examples for object detectors,''
\newblock in {\em 12th USENIX workshop on offensive technologies (WOOT 18)}, 2018.

\bibitem{lee2019physical_adv_patch_det}
Mark Lee and Zico Kolter,
\newblock ``On physical adversarial patches for object detection,''
\newblock {\em arXiv preprint arXiv:1906.11897}, 2019.

\bibitem{thys2019fooling_adv_patch_human_det}
Simen Thys, Wiebe Van~Ranst, and Toon Goedem{\'e},
\newblock ``Fooling automated surveillance cameras: adversarial patches to attack person detection,''
\newblock in {\em Proceedings of the IEEE/CVF conference on computer vision and pattern recognition workshops}, 2019, pp. 0--0.

\bibitem{zhu2021bulb_attack}
Xiaopei Zhu, Xiao Li, Jianmin Li, Zheyao Wang, and Xiaolin Hu,
\newblock ``Fooling thermal infrared pedestrian detectors in real world using small bulbs,''
\newblock in {\em Proceedings of the AAAI Conference on Artificial Intelligence}, 2021, vol.~35, pp. 3616--3624.

\bibitem{zhu2022qr_attack}
Xiaopei Zhu, Zhanhao Hu, Siyuan Huang, Jianmin Li, and Xiaolin Hu,
\newblock ``Infrared invisible clothing: Hiding from infrared detectors at multiple angles in real world,''
\newblock in {\em Proceedings of the IEEE/CVF Conference on Computer Vision and Pattern Recognition}, 2022, pp. 13317--13326.

\bibitem{wei2023hotcold}
Hui Wei, Zhixiang Wang, Xuemei Jia, Yinqiang Zheng, Hao Tang, Shin'ichi Satoh, and Zheng Wang,
\newblock ``Hotcold block: Fooling thermal infrared detectors with a novel wearable design,''
\newblock in {\em Proceedings of the AAAI Conference on Artificial Intelligence}, 2023, vol.~37, pp. 15233--15241.

\bibitem{wei2023shape_attack}
Xingxing Wei, Jie Yu, and Yao Huang,
\newblock ``Physically adversarial infrared patches with learnable shapes and locations,''
\newblock in {\em Proceedings of the IEEE/CVF Conference on Computer Vision and Pattern Recognition}, 2023, pp. 12334--12342.

\bibitem{suard2006pedestrian_infrared}
Fr{\'e}d{\'e}ric Suard, Alain Rakotomamonjy, Abdelaziz Bensrhair, and Alberto Broggi,
\newblock ``Pedestrian detection using infrared images and histograms of oriented gradients,''
\newblock in {\em 2006 IEEE Intelligent Vehicles Symposium}. IEEE, 2006, pp. 206--212.

\bibitem{goodfellow2014explaining}
Ian~J Goodfellow, Jonathon Shlens, and Christian Szegedy,
\newblock ``Explaining and harnessing adversarial examples,''
\newblock {\em arXiv preprint arXiv:1412.6572}, 2014.

\bibitem{ji2021adv_yolo}
Nan Ji, YanFei Feng, Haidong Xie, Xueshuang Xiang, and Naijin Liu,
\newblock ``Adversarial yolo: Defense human detection patch attacks via detecting adversarial patches,''
\newblock {\em arXiv preprint arXiv:2103.08860}, 2021.

\bibitem{müller2021trivialaugment}
Samuel~G. Müller and Frank Hutter,
\newblock ``Trivialaugment: Tuning-free yet state-of-the-art data augmentation,'' 2021.

\bibitem{phan2020resolving_wce}
Trong~Huy Phan and Kazuma Yamamoto,
\newblock ``Resolving class imbalance in object detection with weighted cross entropy losses,''
\newblock {\em arXiv preprint arXiv:2006.01413}, 2020.

\bibitem{FLIR}
``Free - flir thermal dataset for algorithm training | teledyne flir,'' .

\bibitem{s16060820}
Alejandro González, Zhijie Fang, Yainuvis Socarras, Joan Serrat, David Vázquez, Jiaolong Xu, and Antonio~M. López,
\newblock ``Pedestrian detection at day/night time with visible and fir cameras: A comparison,''
\newblock {\em Sensors}, vol. 16, no. 6, 2016.

\bibitem{glenn_jocher_2021_5563715_yolo_v5}
Glenn~Jocher et. al.,
\newblock ``{ultralytics/yolov5: v6.0 - YOLOv5n 'Nano' models, Roboflow integration, TensorFlow export, OpenCV DNN support},'' Oct. 2021.

\bibitem{lin2014microsoft}
Tsung-Yi Lin, Michael Maire, Serge Belongie, James Hays, Pietro Perona, Deva Ramanan, Piotr Doll{\'a}r, and C~Lawrence Zitnick,
\newblock ``Microsoft coco: Common objects in context,''
\newblock in {\em Computer Vision--ECCV 2014: 13th European Conference, Zurich, Switzerland, September 6-12, 2014, Proceedings, Part V 13}. Springer, 2014, pp. 740--755.

\bibitem{rice2020overfitting}
Leslie Rice, Eric Wong, and Zico Kolter,
\newblock ``Overfitting in adversarially robust deep learning,''
\newblock in {\em International Conference on Machine Learning}. PMLR, 2020, pp. 8093--8104.

\bibitem{madry2017towards}
Aleksander Madry, Aleksandar Makelov, Ludwig Schmidt, Dimitris Tsipras, and Adrian Vladu,
\newblock ``Towards deep learning models resistant to adversarial attacks,''
\newblock {\em arXiv preprint arXiv:1706.06083}, 2017.

\bibitem{zhang2019theoretically_trade_off}
Hongyang Zhang, Yaodong Yu, Jiantao Jiao, Eric Xing, Laurent El~Ghaoui, and Michael Jordan,
\newblock ``Theoretically principled trade-off between robustness and accuracy,''
\newblock in {\em International conference on machine learning}. PMLR, 2019, pp. 7472--7482.

\end{thebibliography}

\end{document}